\documentclass[conference]{IEEEtran}
\IEEEoverridecommandlockouts
\usepackage{cite}
\usepackage{amsmath,amssymb,amsfonts}
\usepackage{algorithmic}
\usepackage{graphicx}
\usepackage{textcomp}
\usepackage{xcolor}
\usepackage{multirow}
\usepackage{multicol}

\def\BibTeX{{\rm B\kern-.05em{\sc i\kern-.025em b}\kern-.08em
    T\kern-.1667em\lower.7ex\hbox{E}\kern-.125emX}}
\begin{document}

\title{A Comprehensive Comparison of Machine Learning Based Methods Used in Bengali Question Classification\\
}
\makeatletter
\newcommand{\linebreakand}{%
  \end{@IEEEauthorhalign}
  \hfill\mbox{}\par
  \mbox{}\hfill\begin{@IEEEauthorhalign}
}
\makeatother
%\if 0 
\author{Afra Anika, Md. Hasibur Rahman, Salekul Islam, Abu Shafin Mohammad Mahdee Jameel and \\
Chowdhury Rafeed Rahman \\
\IEEEauthorblockA{\textit{Computer Science and Engineering(CSE)} \\
\textit{United International University(UIU)}\\
Dhaka, Bangladesh \\
\textit{\{aanika161034,mrahman161260\}@bscse.uiu.ac.bd and \{salekul,mahdee,rafeed\}@cse.uiu.ac.bd}
}
}
%\fi
\maketitle

\begin{abstract}
 QA classification system maps questions asked by humans to an appropriate answer category. A sound question classification (QC) system model is the pre-requisite of a sound QA system. This work demonstrates phases of assembling a QA type classification model. We present a comprehensive comparison (performance and computational complexity) among some machine learning based approaches used in QC for Bengali language.    
\end{abstract}


\section{Introduction}
Question classification (QC) deals with question analysis and question labeling  based on the expected answer type. The goal of QC is to assign classes accurately to the questions based on expected answer. In modern system, there are two types of questions \cite{islam2016word}. One is Factoid question which is about providing concise facts and another one is Complex question that has a presupposition which is complex. Question Answering (QA) System is an integral part of our daily life because of the high amount of usage of Internet for information acquisition. 
In recent years, most of the research works related to QA are based on English language such as IBM Watson, Wolfram Alpha. Bengali speakers often fall in difficulty while communicating in English \cite{bhattacharya2001study}.

In this research, we briefly discuss the steps of QA system and compare the performance of seven machine learning based classifiers (Multi-Layer Perceptron (MLP), Naive Bayes Classifier (NBC), Support Vector Machine (SVM), Gradient Boosting Classifier (GBC), Stochastic Gradient Descent (SGD), K Nearest Neighbour (K-NN) and Random Forest (RF)) in classifying Bengali questions to classes based on their anticipated answers. Bengali questions have flexible inquiring ways, so there are many difficulties associated with Bengali QC \cite{islam2016word}. As there is no rich corpus of questions in Bengali Language available, collecting questions is an additional challenge. Different difficulties in building a QA System are mentioned in the literature \cite{hirschman2001natural} \cite{zadeh2006search}. The first work on a machine learning based approach towards Bengali question classification is presented in \cite{islam2016word} that employ the Stochastic Gradient Descent (SGD).  

\section{Related Works}
\subsection{Popular Question-Answering Systems}
Over the years, a handful of QA systems have gained popularity around the world.
 One of the oldest QA system is BASEBALL (created on 1961) \cite{green1961baseball} which answers question related to baseball league in America for a particular season. LUNAR \cite{woods1972lunar} system answers questions about soil samples taken from Apollo lunar exploration. Some of the most popular QA Systems are IBM Watson, Apple Siri and Wolfram Alpha. Examples of some QA systems based on different languages are: Zhang Yu Chinese question classification \cite{yu2005modified} based on Incremental Modified Bayes, Arabic QA system (AQAS) \cite{mohammed1993knowledge} by F. A. Mohammed, K. Nasser, \& H. M. Harb and Syntactic open domain Arabic QA system for factoid questions \cite{fareed2014syntactic} by Fareed et al. QA systems have been built on different analysis methods such as morphological analysis \cite{hovy2000question}, syntactical analysis \cite{zheng2002answerbus}, semantic analysis \cite{wong2007practical} and expected answer Type analysis \cite{benamara2004cooperative}. 

\subsection{Research Works Related to Question Classifications}
Researches on question classification, question taxonomies and QA system have been undertaken in recent years.
There are two types of approaches for question classification according to Banerjee et al in \cite{banerjee2012bengali} - by rules and by machine learning approach. Rule based approaches use some hard coded grammar rules to map the question to an appropriate answer type \cite{prager1999use} \cite{voorhees1999trec}. Machine Learning based approaches have been used by Zhang et al and Md. Aminul Islam et al in \cite{zhang2003question} and \cite{islam2016word}. Many classifiers have been used in machine learning for QC such as Support Vector Machine (SVM) \cite{zhang2003question} \cite{nirob2017question}, Support Vector Machines and Maximum Entropy Model \cite{huang2008question}, Naive Bayes (NB), Kernel Naive Bayes (KNB), Decision Tree (DT) and Rule Induction (RI) \cite{banerjee2012bengali}. In \cite{islam2016word}, they claimed to achieve average  precision of 0.95562 for coarse  class and  0.87646  for finer class using Stochastic Gradient Descent (SGD). 

\subsection{Research Works in Bengali Language }
 A Bengali QC System was built by Somnath Banerjee and Sivaji Bandyopadhyay \cite{banerjee2012bengali} \cite{banerjee2013empirical} \cite{banerjee2013ensemble}. They proposed a two-layer taxonomy classification with 9 coarse-grained classes and 69 fine-grained classes. There are other research works \cite{islam2016word} \cite{kabir2015bangla} in Bengali Language. A survey was performed on text QA techniques \cite{gupta2012survey} where there was an analysis conducted in Bengali Language. Syed Mehedi Hasan Nirob et al achieved 88.62\% accuracy by using 380 top frequent words as the feature in their work \cite{nirob2017question}. 
 
\section{Question Answering (QA) System}
QA system resides within the scope of Computer Science. It deals with information retrieval and natural language processing. Its goal is to automatically answer questions asked by humans in natural language. IR-based QA, Knowledge based approaches and Hybrid approaches are the QA system types. TREC, IBM-Watson, Google are examples of IR-based QA systems. Knowledge based QA systems are Apple Siri, Wolfram Alpha. Examples of Hybrid approach systems are IBM Watson and True Knowledge Evi.   

Figure \ref{fig:QA_sys} provides an overview of QA System. The first step of QA System is \textbf{Question Analysis}. Question analysis has two parts -  \textbf{question classification} and another \textbf{question formulation}. In question classification step, the question is classified using different classifier algorithms. In question formulation, the question is analyzed and the system creates a proper IR question by detecting the entity type of the question to provide a simple answer.

The next step is \textbf{documents retrieval and 
analysis}. In this step, the system matches the query against the sources of answers where the source can be documents or Web. 
In the \textbf{answer extraction} step, the system extracts the answers from the documents of the sources collected in documents retrieval and analysis phase. The extracted answers are filtered and evaluated in \textbf{answer evaluation} phase as there can be multiple possible answers for a query. In the final step, an answer of the question is returned.

\begin{figure}[h]
    \centering
    \includegraphics[width=9cm]{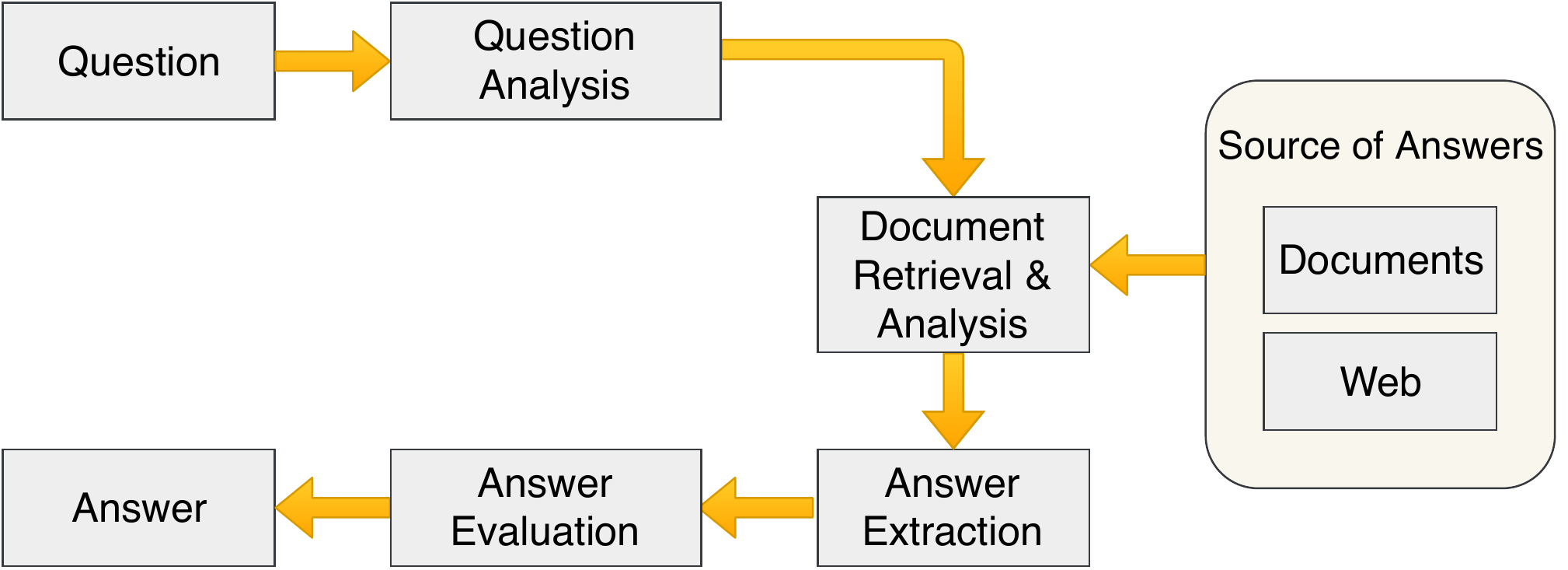}
    \caption{General Architecture of Question Answering System}
\label{fig:QA_sys}
\end{figure}

\section{Proposed Methodology}
 We use different types of classifiers for QA type classification. We separate our methodology into two sections similar to \cite{islam2016word} - one is training section and another is validation section shown in Figure \ref{fig: work_flow}.

\begin{figure}[h]
    \centering
    \includegraphics[width=9cm]{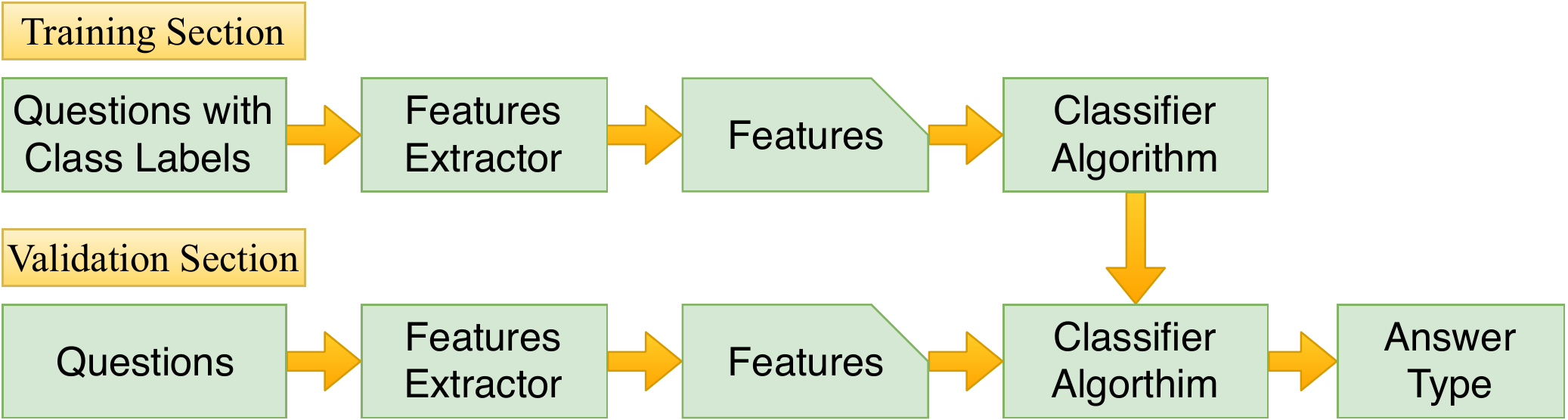}
    \caption{Proposed Work Flow Diagram}
    \label{fig: work_flow}
\end{figure} 

We use 10 fold cross validation where we have 3150 and 350 questions in our training set and validation set respectively. During training, after selecting the possible class labels, the system extracts the features  of the questions and creates a model by passing through a classifier algorithm with the extracted features and class labels. During validation, the system extracts the features of the question and passes it into the model created during training and predicts the answer type.

\section{Question Collection and Categories}
Though Bengali is the seventh most spoken language in terms of number of native speakers \cite{languageref}, there is no standard corpus of questions available \cite{islam2016word}. We have collected total 3500 questions from the Internet and other sources such as books of general knowledge questions, history etc.
The corpus contains the questions and the classes each question belongs to. 

The set of question categories is known as question taxonomy \cite{islam2016word}. We have used two layer taxonomy which was proposed by Xin Li, Dan Roth \cite{li2004semantic}. This two layer taxonomy is made up of two classes which are Coarse Class and Finer Class. There are six coarse classes such as Numeric, Location, Entity, Description, Human and Abbreviation and fifty finer classes such as city, state, mountain, distance, count, definition, group, expression, substance, creative, vehicle etc as shown in the Table I \cite{islam2016word}. A coarse-grained description of a system denotes large components while a fine-grained description denotes smaller sub-components of which the larger ones are composed of.

\begin{table}[h]
\begin{flushleft}
\label{my-label}
\caption{Coarse and fine grained question categories}
\begin{tabular}{|p{2cm}|p{6cm}|}
\hline
\textbf{Coarse Class} & \textbf{Finer Class}\\
\hline
ENTITY (512) & SUBSTANCE (20), SYMBOL (11), CURRENCY (24), TERM (15), WORD (20), LANGUAGE (30), COLOR (15), RELIGION (15), SPORT (10), BODY (10), FOOD (11), TECHNIQUE (10), PRODUCT (10), DISEASE (10), OTHER (22), LETTER (10), VEHICLE (11), PLANT (12), CREATIVE (216), INSTRUMENT (10), ANIMAL (10), EVENT (10)\\
\hline
NUMERIC (889) & COUNT (213), DISTANCE(13),CODE(10), TEMPERATURE (13), WEIGHT (20), MONEY (10), PERCENT (27), PERIOD (33), OTHER (34), DATE (452), SPEED (10), SIZE (54)\\
\hline
HUMAN (669) & INDIVIDUAL (618), GROUP (18), DESCRIPTION (23), TITLE (10) \\
\hline
LOCATION (650) & MOUNTAIN (32), COUNTRY (125), STATE (98), OTHER (121), CITY (274) \\
\hline
DESCRIPTION (248) & DEFINITION (153), REASON (44), MANNER (22), DESCRIPTION (39) \\
\hline
ABBREVIATION (532) & ABBREVIATION (519), EXPRESSION (13) \\
\hline
\end{tabular}
\end{flushleft}
\label{Table:dataset}
\end{table}

\section{Implementation of the System}
\subsection{Feature Extraction}
Question word answer phrases, parts of speech tags, parse feature, named entity and semantically related words are different features from answer type detection \cite{huang2008question}. We use question word and phrases as features for answer type detection. We consider the following features:

\subsubsection{TF-IDF}
Term Frequency - Inverse Document Frequency (TF-IDF) is a popular method used to identify the importance of a word in a particular document. TF-IDF transforms text into meaningful numeric representation. This technique is widely used to extract features for Natural Language Processing (NLP) applications \cite{ramos2003using} \cite{hakim2014automated}. 

\subsubsection{Word level N-Grams}
N-grams is n-back to back words in a text. Queries of a same class usually share word n-grams \cite{islam2016word}. In this system, we choose bi-gram for extracting features. 

\subsubsection{Stop Words}
We use two setups (as done in \cite{islam2016word}) for our system. In the first setup, we eliminate the stop words from the text using another dataset containing only stop words. At second step, we work without eliminating the stop words from the text which gives better result than the first setup. 

\subsection{Classification Algorithms}

\subsubsection{Multi Layer Perceptron (MLP)}
MLP contains three layers - an input layer, an output layer and some hidden layers. Input layer receives the signal, the output layer gives a decision or prediction about the input and the computation of the MLP is conducted in the hidden layers. In our system, we use 100 layers. For weight optimization, we use Limited-memory Broyden–Fletcher–Goldfarb–Shanno (LBFGS) optimization algorithm.

\subsubsection{Support Vector Machine (SVM)}
SVM gives an optimal hyper-plane and it maximizes the margin between classes. We use Radial Basis Function (RBF) kernel in our system to make decision boundary curve-shaped. For decision function shape, we use the original one-vs-one (ovo) decision function. 

\subsubsection{Naive Bayesian Classifier (NBC)}
NBC is based on Bayes' Theorem which gives probability of an event occurrence based on some conditions related to that event. We use Multinomial Naive Bayes Classifier with smoothing parameter equals to 0.1. A zero probability cancels the effects of all the other probabilities.

\subsubsection{Stochastic Gradient Descent (SGD)}
Stochastic gradient descent optimizes an objective function with suitable smoothness properties \cite{hardt2015train}. It selects few examples randomly instead of whole data for each iteration. We use 'L2' regularization for reduction of overfitting.

\subsubsection{Gradient Boosting Classifier (GBC)}
Gradient Boosting Classifier produces a prediction model consisting of weak prediction models. Gradient boosting uses decision trees. We use 100 boosting stages in this work.

\subsubsection{K Nearest Neighbour (K-NN)}
K-NN is a supervised classification and regression algorithm. It uses the neighbours of the given sample to identify its class. K determines the number of neighbours needed to be considered. We set the value of K equals to 13 in this work.

\subsubsection{Random Forest (RF)}
RF is an ensemble learning technique. It constructs large number of decision trees during training and then predicts the majority class. We use 500 decision trees in the forest and "entropy" function to measure the quality of a split.

\subsection{Results and Discussion}
 Table II shows the \textbf{accuracy} and \textbf{F1 score} for different classifiers with and without eliminating stop words while extracting features. Figure \ref{fig:bar_chart} shows the average results of different classifiers in a bar chart with and without eliminating stop words from the questions.

Overall, SGD has shown the best performance on our dataset as it introduces non-linearity and uses back-propagation for updating parameter weights using loss function calculated on training set into classification. K-NN has shown the weakest performance overall, as this algorithm has a bad reputation of not working well in high dimensional data \cite{pestov2013k}. MLP and SVM have shown similar performance. MLP takes advantage of multiple hidden layers in order to take non-linearly separable samples in a linearly separable condition. SVM accomplishes this same feat by taking the samples to a higher dimensional hyperplane where the samples are linearly separable. Gradient Boosting Classifier (GBC) and Random Forest (RF) both utilize a set of decision trees and achieve similar results (RF performs slightly without eliminating stop words). Naive Bayesian Classifier (NBC) shows performance on per with GBC and RF algorithms. The overall better performance of all the algorithms when provided with stop words show the importance of stop words in Bengali QA classification. 
\begin{table}[]
\centering
\caption{Experiment Results}
\begin{tabular}{|c|l|l|l|l|l|l|l|}
\hline
\multicolumn{8}{|c|}{After Eliminating Stop Words} \\ 
\hline
& MLP   & SVM   & NBC   & SGD   & GBC   & KNN   & RF    \\ \hline
Accuracy & 0.779 & 0.741 & 0.724 & 0.797 & 0.701 & 0.376 & 0.712 \\ 
\hline
F1 Score & 0.761 & 0.705 & 0.693 & 0.775 & 0.686 & 0.439 & 0.680 \\ 
\hline
\multicolumn{8}{|c|}{Without Eliminating Stop Words}    \\ \hline
& MLP   & SVM   & NBC   & SGD   & GBC   & KNN   & RF    \\ \hline
Accuracy & 0.83  & 0.801 & 0.789 & 0.832 & 0.792 & 0.781 & 0.816 \\ 
\hline
F1 Score & 0.810 & 0.765 & 0.759 & 0.808 & 0.775 & 0.755 & 0.783 \\ 
\hline
\end{tabular}
\end{table}

\begin{figure}[h]
    \centering
    \includegraphics[width=9cm]{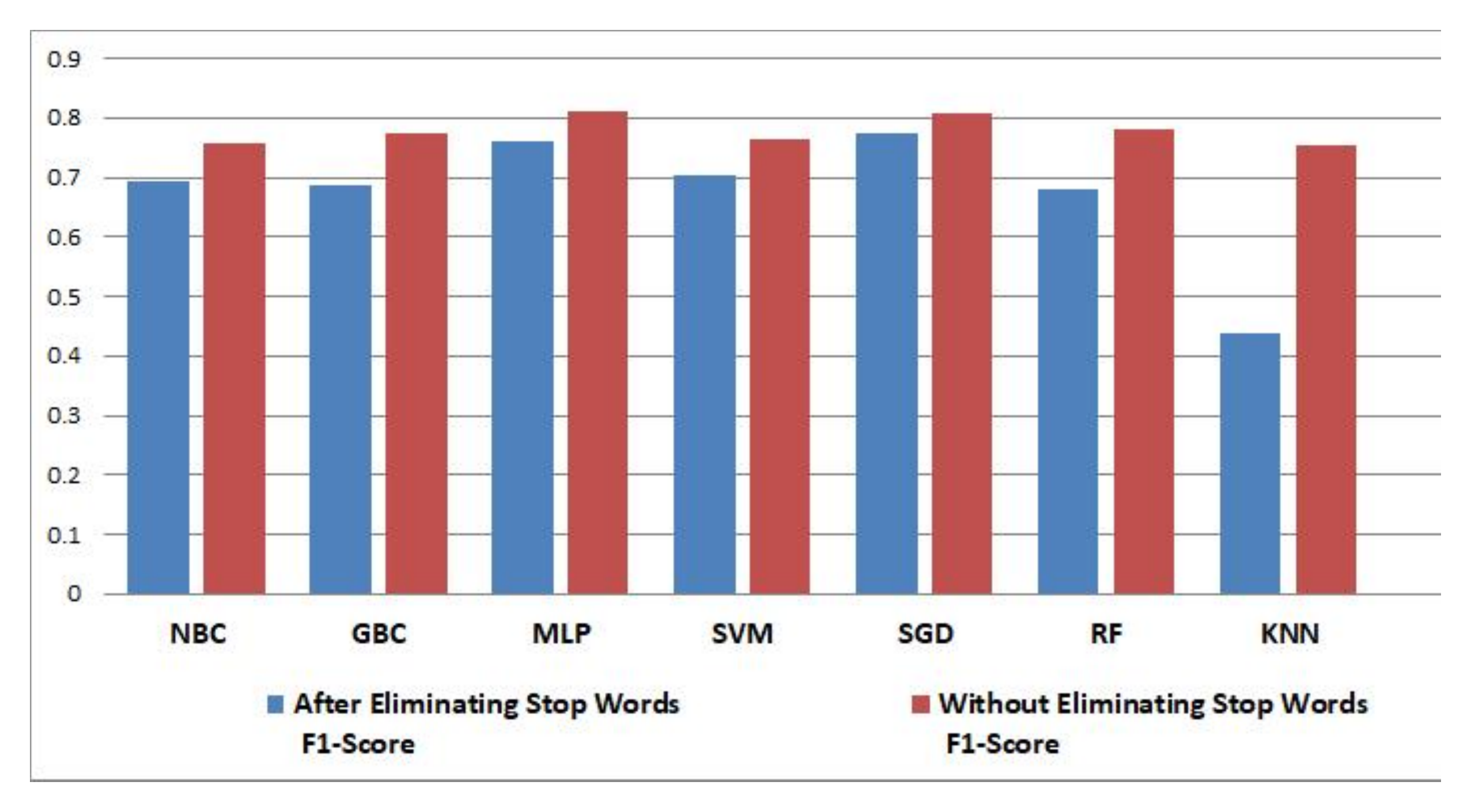}
    \caption{F1 Scores of Each Classifiers}
    \label{fig:bar_chart}
\end{figure}

Figure \ref{fig:QA} shows the predictions of some particular questions by each of the classifiers. The input is a full question and the output is the class of the question. 

\begin{figure}[h]
    \begin{flushleft}
    \includegraphics[width=9.2cm]{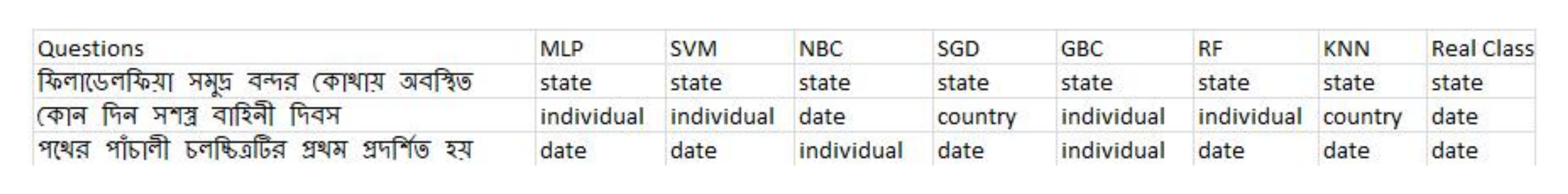}
    \caption{Prediction of Questions}
    \label{fig:QA}
    \end{flushleft}
\end{figure}

\subsection{Computational Complexity}
\begin{table}[h]
\label{my-label}
\caption{Computational Complexity of Each Classifier}
\begin{tabular}{|p{2cm}|p{2.8cm}|p{2.8cm}|}
\hline
   &       Training Section   &    Prediction Section    \\
    \hline
MLP &O(nph\textsuperscript{k}i)&O(nph\textsuperscript{k}) \\ 
\hline
SGD &   O(in$\overline{m}$)     &                      \\ 
\hline
SVM &O(n\textsuperscript{2}p+n\textsuperscript{3})&O(n\textsubscript{sv}p)            \\ 
\hline
NBC &       O(np)                                 &           O(p)             \\ 
\hline
GBC &  O(npn\textsubscript{trees})&O(pn\textsubscript{trees})\\ 
\hline
kNN &           &             O(np)          \\ 
\hline
RF  &O(n\textsuperscript{2}pn\textsubscript{trees})&O(pn\textsubscript{trees})  \\ 
\hline
\end{tabular}
\label{Table : Complexity}
\end{table}
In Table \ref{Table : Complexity}, n = No. of training sample, p = No. of features, n\textsubscript{trees} = No. of trees (for methods based on various trees), n\textsubscript{sv} = No. of support vectors, i = No. of iterations, h = No. of nodes in each hidden layer, k = No. of hidden layers and $\overline{m}$ = the average no. of non-zero attributes per sample.

\section{Conclusion}
 By implementing different machine learning based classifiers on our Bengali question corpus, we perform a comparative analysis among them. The question classification impacts the QA system. So, it is important to classify the question more precisely. This work will help the research community to choose a proper classification model for smart Bengali QA system development. Future work should aim at developing a richer corpus of Bengali questions which will help in getting better vector representation of words and thus will facilitate deep learning based automatic feature extraction.

\bibliography{main}
\bibliographystyle{plain}

\end{document}